\documentclass{article}
\pdfoutput=1

\usepackage[numbers]{natbib}
\usepackage{arxiv}

\usepackage[utf8]{inputenc}
\usepackage[T1]{fontenc}
\usepackage{hyperref}
\usepackage{url}
\usepackage{booktabs}
\usepackage{amsfonts}
\usepackage{nicefrac}
\usepackage{microtype}

\usepackage{times}
\usepackage{graphicx}
\usepackage{amsmath}
\usepackage{amssymb}
\usepackage{tikz}
\usetikzlibrary{
  arrows.meta,
  fit,
  positioning,
  patterns
}
\usepackage{listings}
\usepackage{bm}
\usepackage{subcaption}

\tikzset{
  neuron/.style={
    circle,draw,thick,
    inner sep=0pt,
    minimum size=1.em,
    node distance=1ex and 2em,
    fill=gray!20,
  },
  group/.style={
    rectangle,draw,thick,
    inner sep=5pt,
    rounded corners = 3pt,
  },
  groupx/.style={
      group,
      label=above:\textbf{x},
  },
  groupy/.style={
      group,
      label=above:\textbf{y},
  },
  h1/.style={
      label=above:$\bm{h_1}$,
  },
  h2/.style={
      label=above:$\bm{h_2}$,
  },
  io/.style={
    neuron,
    fill=gray!30,
  },
  conn/.style={
    ->,
    >=stealth,
    thick,
    anchor=south,
    shorten >=.1cm,
    shorten <=.1cm,
  },
  connsym/.style={
    conn,
    <->
  },
  connback/.style={
    conn,
    transform canvas={yshift=-.5cm}
  },
  weight/.style={
    anchor=south,
  },
  weightback/.style={
    weight,
    anchor=north,
  },
}

\newcommand{\DD}[2]{\frac{\mathrm{d} #1}{\mathrm{d} #2}}
\newcommand{\DDT}[1]{\DD{#1}{t}}

\newenvironment{cenumerate}
{ \begin{enumerate}
    \setlength{\itemsep}{0pt}
    \setlength{\parskip}{0pt}
    \setlength{\parsep}{0pt}     }
{ \end{enumerate}                  }
\renewcommand{\vec}[1]{\boldsymbol{#1}}

\graphicspath{ {./ffe_images/} }

\begin{document}

\title{Error Forward-Propagation: Reusing Feedforward Connections to Propagate Errors in Deep Learning}

\author{Adam A. Kohan\\
University of Massachusetts Amherst\\
{\tt\small akohan@cs.umass.edu}
\And
Edward A. Rietman\\
University of Massachusetts Amherst\\
{\tt\small erietman@umass.edu}
\And
Hava T. Siegelmann\\
University of Massachusetts Amherst\\
{\tt\small hava@cs.umass.edu}
}

\maketitle

\begin{abstract}
    We introduce Error Forward-Propagation, a biologically plausible mechanism to propagate error feedback forward through the network. Architectural constraints on connectivity are virtually eliminated for error feedback in the brain; systematic backward connectivity is not used or needed to deliver error feedback. Feedback as a means of assigning credit to neurons earlier in the forward pathway for their contribution to the final output is thought to be used in learning in the brain. How the brain solves the credit assignment problem is unclear. In machine learning, error backpropagation is a highly successful mechanism for credit assignment in deep multilayered networks. Backpropagation requires symmetric reciprocal connectivity for every neuron. From a biological perspective, there is no evidence of such an architectural constraint, which makes backpropagation implausible for learning in the brain. This architectural constraint is reduced with the use of random feedback weights. Models using random feedback weights require backward connectivity patterns for every neuron, but avoid symmetric weights and reciprocal connections. In this paper, we practically remove this architectural constraint, requiring only a backward loop connection for effective error feedback. We propose reusing the forward connections to deliver the error feedback by feeding the outputs into the input receiving layer. This mechanism, Error Forward-Propagation, is a plausible basis for how error feedback occurs deep in the brain independent of and yet in support of the functionality underlying intricate network architectures. We show experimentally that recurrent neural networks with two and three hidden layers can be trained using Error Forward-Propagation on the MNIST and Fashion MNIST datasets, achieving $1.90\%$ and $11\%$ generalization errors respectively.
\end{abstract}

\section{Introduction}
Error backpropagation has been the primary workhorse of learning artificial neural networks and often attributed as behind the success of deep learning. In deep learning, neural networks with deep architectures \cite{ba2014deep} of non-linear rate based neurons connected by synaptic weights are activated layer-by-layer starting with a given input to compute a prediction. Deep learning algorithms are designed to exploit the deeper layers in multilayered architectures, called hidden layers, to choose the weights (and other parameters) in order to learn complex, hierarchical representations \cite{lecun2015deep}. In a supervised setting, neural networks learn the target of the input, given the prediction and target. Typically, the prediction is compared with the target as the training objective and the difference forms an error signal. The error signal is propagated backward through the neural network in reverse order of forward layer-wise activations. The weights are updated depending on the layer-wise activations, but also on the credit assigned for the error \cite{rumelhart1986learning}. Backpropagation is used to calculate the direction of weight updates in hidden layers. Learning is non-local and has no notion of physical time.

Neurons in artificial neural networks are inspired by neurons in the brain. The computation in each neuron in the network resembles those in biological neurons. However, learning with standard backpropagation and learning in neuroscience are currently difficult to reconcile \cite{marblestone2016toward}.

Biological neural networks are composed of spiking neurons with continuous time dynamics and non-linear forward activation. Learning is believed to be local in the brain, dependent on physical time, and seems to lack the comprehensive reciprocal or bidirectional connectivity and weight symmetry fundamental to backpropagation. Backpropagation requires buffering the successive forward layer-wise activations and weights from the hidden node to the source of the error until the error signal arrives to assign credit and calculate the hidden weight update. This difference between biology and backpropagation has been identified as an issue for a long while and been named the weight transport problem \cite{grossberg1987competitive}. Still, error signals, local or global, from differences in predictions and targets are thought to be used in learning in the brain. The required mechanism underlying the propagation of such error signals and their application in adjusting weights is not clear and under ongoing investigation.

Only recently have biological plausible proposals been made to adapt backpropagation to spiking and continuous time neural networks with reasonable success. Local contrastive Hebbian learning in energy based models are one class of proposals that have shown a level of equivalence with recurrent backpropagation \cite{scellier2017equivalence,xie2003equivalence, scellier2017equilibrium}. In these learning algorithms, the networks minimizes the difference between the fixed points when running freely and when clamped to an external input for error correction. Symmetric and, as recently reported, random feedback weights work for these algorithms \cite{scellier2018extending}. The class of proposals based on random feedback weights \cite{liao2016important,guerguiev2017towards,lillicrap2016random,nokland2016direct,neftci2017event} avoid weight symmetry. It was shown that the sign concordance between the forward and feedback weights is enough to deliver effective error signals. The forward weights align with the random feedback weights and achieve functionally equivalent approximate symmetry. Random feedback weights are also sufficient to deliver feedback for calculating error locally. These works have made breakthroughs in the possible mechanics of credit assignment taking place in the brain.

Even so, the models based on these proposals and backpropagation depend on a problematic assumption. The architectures of models based on backpropagation rely on systematic feedback connections to layers and neurons. Though it is possible \cite{lillicrap2016random,scellier2017equilibrium,guerguiev2017towards}, there is no evidence of such complete structural enforcement in the neocortex. The brain does not seem to exhibit the comprehensive level of feedback connectivity necessary for every neuron to receive feedback, even if the connectivity is not symmetric, reciprocal, or direct.

\begin{figure}
    \centering
    \begin{subfigure}[b]{\textwidth}
        \centering
        \begin{tikzpicture}
          \node[neuron] at (2.4,1) (n1a) {};
          \node[neuron,below=of n1a] (n1b) {};
          \node[neuron,below=of n1b] (n1c) {};
          \node[group,h1,fit={(n1a) (n1b) (n1c)}] (l1) {};

          \node[neuron] at (4.8,1) (n3a) {};
          \node[neuron,below=of n3a] (n3b) {};
          \node[neuron,below=of n3b] (n3c) {};
          \node[group,h2,fit={(n3a) (n3b) (n3c)}] (l3) {};
          \draw[conn] (l1) -- node[weight] {$W_{12}$} (l3);

          \node[io] at (7.2,1) (y1) {};
          \node[io,below=of y1] (y2) {};
          \node[io,below=of y2] (y3) {};
          \node[groupy,fit={(y1) (y2) (y3)}] (yt) {};
          \draw[conn] (l3) -- node[weight] {$W_{23}$} (yt);
          \draw[conn] (yt) to [out=180+45,in=-45,looseness=1] node[weight,anchor=north] {$W_{31}$} (l1);

          \node[io] at (0,1) (x1) {};
          \node[io,below=of x1] (x2) {};
          \node[io,below=of x2] (x3) {};
          \node[groupx,fit={(x1) (x2) (x3)}] (xt) {};
          \draw[conn] (xt) -- node[weight] {$W_{01}$} (l1);
        \end{tikzpicture}
        \caption{Error Forward-Propagation}
        \label{fig:archefp}
        \vspace{.5cm}
    \end{subfigure}
    \begin{subfigure}[b]{\textwidth}
        \centering
        \begin{tikzpicture}
            \node[neuron] at (2.4,1) (n1a) {};
          \node[neuron,below=of n1a] (n1b) {};
          \node[neuron,below=of n1b] (n1c) {};
          \node[group,h1,fit={(n1a) (n1b) (n1c)}] (l1) {};

          \node[neuron] at (4.8,1) (n3a) {};
          \node[neuron,below=of n3a] (n3b) {};
          \node[neuron,below=of n3b] (n3c) {};
          \node[group,h2,fit={(n3a) (n3b) (n3c)}] (l3) {};
          \draw[conn] (l1) -- node[weight] {$W_{12}$} (l3);
          \draw[connback] (l3) -- node[weightback] {$W_{21}$} (l1);

          \node[io] at (7.2,1) (y1) {};
          \node[io,below=of y1] (y2) {};
          \node[io,below=of y2] (y3) {};
          \node[groupy,fit={(y1) (y2) (y3)}] (yt) {};
          \draw[conn] (l3) -- node[weight] {$W_{23}$} (yt);
          \draw[connback] (yt) -- node[weightback] {$W_{32}$} (l3);

          \node[io] at (0,1) (x1) {};
          \node[io,below=of x1] (x2) {};
          \node[io,below=of x2] (x3) {};
          \node[groupx,fit={(x1) (x2) (x3)}] (xt) {};
          \draw[conn] (xt) -- node[weight] {$W_{01}$} (l1);
          \draw[connback] (l1) -- node[weightback] {$W_{10}$} (xt);

        \end{tikzpicture}
        \caption{Random Feedback Weights}
        \label{fig:archrfw}
        \vspace{.5cm}
    \end{subfigure}
    \begin{subfigure}[b]{\textwidth}
        \centering
        \begin{tikzpicture}
            \node[neuron] at (2.4,1) (n1a) {};
          \node[neuron,below=of n1a] (n1b) {};
          \node[neuron,below=of n1b] (n1c) {};
          \node[group,h1,fit={(n1a) (n1b) (n1c)}] (l1) {};

          \node[neuron] at (4.8,1) (n3a) {};
          \node[neuron,below=of n3a] (n3b) {};
          \node[neuron,below=of n3b] (n3c) {};
          \node[group,h2,fit={(n3a) (n3b) (n3c)}] (l3) {};
          \draw[connsym] (l1) -- node[weight] {$W_{12}$} (l3);

          \node[io] at (7.2,1) (y1) {};
          \node[io,below=of y1] (y2) {};
          \node[io,below=of y2] (y3) {};
          \node[groupy,fit={(y1) (y2) (y3)}] (yt) {};
          \draw[connsym] (l3) -- node[weight] {$W_{23}$} (yt);

          \node[io] at (0,1) (x1) {};
          \node[io,below=of x1] (x2) {};
          \node[io,below=of x2] (x3) {};
          \node[groupx,fit={(x1) (x2) (x3)}] (xt) {};
          \draw[connsym] (xt) -- node[weight] {$W_{01}$} (l1);

        \end{tikzpicture}
        \caption{Symmetric Feedback Weights}
        \label{fig:archsym}
    \end{subfigure}
    \caption{(a) Error Forward-Propagation reuses the feedforward connections by delivering the error or feedback via loop connections. (b) Random Feedback Weights deliver a random projection of the error or feedback to neurons in the hidden layers. (c) Symmetric Weights delivers the error by multiplying with the transpose of the forward weight matrix.}
    \label{fig:archcompare}
\end{figure}
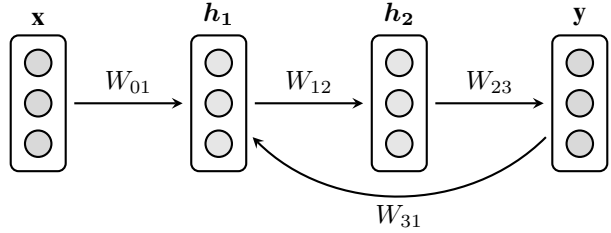
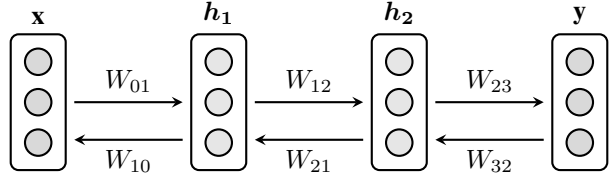
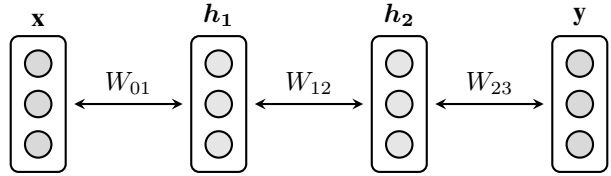

Here, we show that by feeding the output layer into the input receiving layer ($h1$ in figure \ref{fig:archefp}), feedforward connections can be reused to deliver effective feedback, much the same as symmetric or random feedback weights can. Specifically, feedforward connection can be reused to move hidden neuron activity toward its' target firing rate quickly. We refer to this mechanism as Error Forward-Propagation.

Error Forward-Propagation is based on our observations into the core behavior of biologically plausible learning algorithms based on backpropagation, particularly in the context of advances made in \cite{scellier2017equilibrium,lillicrap2016random,guerguiev2017towards} relaying feedback for local errors. We found that at the core of these algorithms is the ability of the network to express two states in a hidden neuron at different time scales. The hidden neurons' current prediction firing rate and target firing rate are expressed at two sufficiently distant time scales such that updating the weights to close the difference between the two states pushes the output neurons' prediction towards the target. We propose that Error Forward-Propagation is able to help facilitate this behavior as well symmetric or random feedback weights can for deep learning.

This simple mechanism combined with the weight update portion of a backpropagation like algorithm avoids the need for comprehensive feedback connectivity to propagate the feedback to hidden neurons. However, the learning algorithm needs to be capable of encoding the separation between the prediction firing rate and the target firing rate. Many learning algorithms, especially backpropagation like algorithms, have two phases of learning which is sufficient to separate the two firing rates. Alternatively, a single phase learning algorithm with two different timescales (filters) for each neuron may be able to differentiate between the two firing rates at the point that they switch from one to the other.

Error Forward-Propagation is a basic mechanism. It is compatible with more simple or complex learning algorithms than the one used here. It is suitable to more sophisticated network architectures, especially those with deep feedforward only regions. It also does not assume backward connectivity is for error feedback, leaving their function open to other possibilities. Our findings drastically reduce the architectural constraints on quick and accurate deep error propagation.

\section{Neural Network Model}
We train deep recurrent networks with a neuron model based on the continuous Hopfield model \cite{hopfield1984neurons}:
\begin{align} \label{eq:neuronmodel}
    \DDT{s_j} &= \DD{\rho(s_j)}{s_j} (\sum_{i \to j} w_{ij} \rho(s_i) + \sum_{i \in I \to j \in O} w_{ij} \rho(s_i) + b_j)\\\nonumber
            &- \frac{s_j}{r_j} - \beta \sum_{j \in O} {(s_j - d_j)}
\end{align}
where $s_j$ is the state of neuron $j$, $\rho(s_j)$ is a non-linear monotone increasing function of it's firing rate, $b_j$ is the bias, $\beta$ limits magnitude and direction of the feedback, $O$ is the subset of output neurons, $I$ is the subset of input receiving neurons, and $d_j$ is the target for output neuron $j$. The input receiving neurons, $s_j in I$, are the neurons with forward connections from the input layer.

The networks are entirely feedforward except for the final feedback loop from the output neurons $s_j \in O$ to the input receiving neurons $s_j \in I$. All weights and biases are trained. The weights in the feedback loop connections may be fixed or trained. The output neurons receive the $L_2$ error as an additional input which nudges the firing rate towards the target firing rate $d_j$. The target firing rate $d_j$ is the one-hot vector of the target value; all tasks in this paper are classification tasks. 

\section{Learning Algorithm}
Biological neural networks work in continuous time and have no indication of different dynamics in prediction and learning. The learning framework, Equilibrium Propagation (EP), proposed in \cite{scellier2017equilibrium} is one way to introduce physical time in deep learning and have the same dynamics in prediction and learning. Only one type of computation can be used for both prediction and learning, avoiding the need for different hardware for each \cite{scellier2017equilibrium}. The learning algorithm underlying the framework can train the class of energy based models in which the continuous Hopfield model falls into and our network model is based on. However, the requirement of an energy function can be safely relaxed, such as the need for symmetric weights \cite{scellier2018extending}, and as we will see, the need for almost all error feedback connections. We use a new learning algorithm for Error Forward-Propagation based on the learning algorithm from Equilibrium Propagation described in \cite{scellier2017equilibrium}.

\subsection{Equilibrium Propagation}
Equilibrium Propagation combines a local contrastive Hebbian rule (CHL) with ideas of recurrent backpropagation for recurrent networks of neurons with graded responses. The network dynamics are defined by taking the negative gradient of an energy function. The prototypical energy function is a kind of Hopfield energy \cite{hopfield1984neurons}
\begin{align} \label{eq:prototypical}
    E &= \sum_{j} \frac{1}{r_j} s_j^2
        -\frac{1}{2} \sum_{i \neq j} w_{ij} \rho(s_i)\rho(s_j) - \sum_{j} b_j \rho(s_j)
\end{align}
where $s_j$ is the state of neuron $j$, $\rho(s_j)$ is a non-linear monotone increasing function of it's firing rate, and $b_j$ is the bias, with symmetric weights $w_{ij} = w_{ji}$. Note that weights $w_{ij}$ may be set to $0$ for designing different network architectures that are not all-to-all, such as layered networks.
The cost function is defined as
\begin{align}
    C &= \sum_{j \in O} \frac{1}{2} {||s_j - d_j||}^2
\end{align}
where $O$ is the subset of output neurons and $d_j$ is the target for output neuron $j$. The total energy function is then
\begin{align}
    F &= E + \beta C
\end{align}
where $\beta$ is the clamping factor, explained further below.

The EP learning algorithm can be broken into the forward phase, the clamped phase, and the update rule. In the forward phase, the input neurons are fixed to a given value and the network is relaxed to an energy minimum to produce a prediction. In the clamped phase, the input neurons remain fixed and the rate of output neurons $s_j \in O$ are perturbed toward the target value $d_j$, given the prediction $s_j$, which propagates to connected hidden layers. The update rule is a simple contrastive Hebbian plasticity mechanism that subtracts $s_i^0 s_j^0$ at the energy minimum (fixed point) in the forward phase from $s_i^\beta s_j^\beta$ after the perturbation of the output, when $\beta > 0$:

\begin{align}\label{eq:equpdate}
    \Delta W_{ij} &\propto \frac{1}{\beta} ( \rho(s_i^\beta) \rho(s_j^\beta) -  \rho(s_i^0) \rho(s_j^0))
\end{align}

The clamping factor $\beta$ allows the network to be sensitive to internal perturbations. As $\beta \to +\infty$, the fully clamped state in general CHL algorithms is reached where perturbations from the objective function tend to overrun the dynamics and continue backwards through the network. The objective function is local to $\beta$ and not determined by the energy function $E$ as in general CHL. The clamping factor and local objective function mean the energy state stays close to the relaxed energy minimum and there is more flexibility in choice of cost $C$ and energy $E$ functions.

\subsection{Feedforward Propagation}
We introduce a new learning algorithm based on the Equilibrium Potential learning algorithm. The network model in EP has complete reciprocal connections $\forall s_i s_j: s_j s_i$ for propagating the error backwards through the network, equation \ref{eq:prototypical} and figure \ref{fig:archsym}. Feedforward Propagation (FP) has a less restrictive architecture which does not have reciprocal connections, but instead has feedback loop connections ${\forall s_j \in O}, {\forall s_i \in I}: s_j s_i$ for propagating the error forward through the network, equation \ref{eq:neuronmodel} and figure \ref{fig:archefp}. The learning algorithm in EP is designed for unidirectional connections and is slow due to the length relaxation phase. Feedforward Propagation learns in the same way as Equilibrium Potential, but has an update rule for unidirectional connections and improves the forward and clamped phases to shorten the relation phase immensely.

\subsubsection{Update Rule}
This update rule in the EP learning algorithm is an approximation of $\DD{}{\beta} (\rho(s_i)\rho(s_j)) \approx (\rho(s_i^\beta) \rho(s_j^\beta) -  \rho(s_i^0) \rho(s_j^0))$. (Refer to \cite{scellier2017equilibrium} for the complete derivation of the original update rule and learning algorithm.) The approximation updates the parameters taking into consideration the difference between the prediction and target firing rates for both the presynaptic and postsynaptic neurons. The approximation seems to fit the EP energy-based model well because it has reciprocal connections and asymmetric weights. However, with Error Forward-Propagation, the network only has feedback loops, from the output neurons to the input receiving neurons, which are treated as feedforward connections. The network makes predictions in only one direction, forward.

For a synapse (connection) and keeping the training example fixed, the presynaptic neuron will relax before the postsynaptic neuron or stop impacting the postsynaptic neuron before the postsynaptic neuron relaxes. Remember that the learning algorithm updates the parameters once the network has relaxed to the fixed points in forward and clamped phases. Since the inputs are fixed to a training example, they are constant before and at the point that the input receiving neurons relax and the outgoing synapses (weights) are updated. Similarly, since the presynaptic neurons effectively relax before the postsynaptic neurons, they can be considered as constant before and at the point that the postsynaptic neurons relax and the synapses are updated.

This behavior can be encoded in the update rule. For the update, the presynaptic neuron can be held constant and a better approximation for unidirectional connections during inference is
\begin{align} \label{eq:uniupdate}
    \Delta W_{ij} &\propto \rho(s_i) \DD{}{\beta} (\rho(s_j)) \\\nonumber
       & \approx \frac{1}{\beta} \rho(s_i^0) (\rho(s_j^\beta) - \rho(s_j^0))
\end{align}
This is update rule used for Feedforward Propagation.

\subsubsection{Forward and Clamped Phases}
The network parameters are treated as constants during the forward and clamped phases in the EP learning algorithm. The forward phase, clamped phase, and update rule are run once (possibly in two parts for the update rule) per epoch per training example. However, the parameters of the network are not considered as constants in the energy function. They are differentiable variables and their movement is expected in the relaxation of the network to a fixed point.

We propose that if and how frequently network parameters are updated in sequence with the free and clamped phases effects the length of relaxation period in the free phase. The free phase, and clamped phase, update rule are looped over as part of a larger phase. In this scheme, the network parameters are no longer held constant, but are also not updated continuously as the neurons' state updates are. This is to allow the network to start settling and be moving toward the fixed point when the network parameters are updated. The implementation in Feedforward Propagation is as follows:

\begin{lstlisting}
for x in examples:
    for i in range(0, num_loops):
        free_phase(x, num_free)
        weakly_clamped_phased(x, num_clamped)
        update_rule(x, parameters)
\end{lstlisting}

\section{Implementation of the Model}
Our model is a recurrent rate-based feedforward neural network with a feedback loop from the output neurons to the input receiving neurons. The implementation is of layered networks with 2-3 hidden layers. More complex architectures are possible in our implementation, but are not considered here. The model was implemented in Python and Theano \cite{bergstra2010theano,bastien2012theano}.

The differential equation of the neuron is implemented using the Euler method:
\begin{align} \label{eq:diffeuler}
    s_j \leftarrow s_j + \min(1,\max(0, h\DDT{s_j}(\theta,x,\beta,s)))
\end{align}
where $h$ is the step size ans $\theta$ is the parameters. The clipping of the third term is do to the choice of activation function \cite{scellier2017equilibrium}. The hard sigmoid activation function $\rho(s_j) = \min(1,\max(0,s_j))$ keeps the neuron within $[0,1]$ since $\rho'(s_j) = 0$ otherwise. The neuron's motion is estimated and as such may not necessarily stay within those bounds. For example, a set of incoming potentials may be large enough to overwhelm the step size and push the neuron past the bounds. To prevent the neuron from remaining forever outside of $[0,1]$ in such cases, the state update is clipped. The step size $h$ is selected to be a fraction (divisible and multiple) of the time constant $d_j$. We choose $h = .5$, the same as in \cite{scellier2017equilibrium}.

The datasets are split into training, validation, and test sets. The targets of the dataset are one-hot encoded $d_j$ for learning and validating. Training proceeds as in the learning algorithm for each example $x$ and target $d$:
\begin{cenumerate}
    \item clamp the input neurons to $x$
    \item run the free phase for a multiple of the length of the network and store the state of the network.
    \item run the weakly clamped phases for a smaller multiple of the length of the network with $\beta \ge 0$
    \item update the parameters according to \ref{eq:uniupdate}
    \begin{align} \label{eq:update}
        \Delta W_{ij} &\propto \frac{1}{\beta} \rho(s_i^0) (\rho(s_j^\beta) - \rho(s_j^0)) \nonumber
    \end{align}
    \item repeat the last three steps for a small number of iterations
\end{cenumerate}

The prediction is made after the last iteration of the free phase, but before the weakly clamped phase. The predicted value $y_{pred}$ is selected from the output neuron with the maximum firing rate, and the error is calculated from the same neuron.
\begin{align}
    y_{pred} &= \text{argmax}_{i} \left(  \right(s_j)_i)
\end{align}

The duration of the free and clamped phases should be at least long enough to propagate a input across the network. The time needed for a neuron to settle to an input value is it's time constant $r_j$. The time needed to propagate an input across the network is a multiple of the networks length $L$ and the time needed for a single neuron settle to an input. So the free and clamped phases should be $c (L)(r_j)$ long or $c (L) (r_j / h)$ iterations. With the FP learning algorithm, we observe that the duration of the free phase should be short, a small $c$. We find that two to four times the length of the network works well. The clamped phase can be even shorter, only the length of the network, as observed in \cite{scellier2017equilibrium} and confirmed here. The number of iterations needed of the free phase, clamped phase, and update rule is found to be small in our experiments. Two to three iterations works well for networks with one to four hidden layers.

\begin{table*}
\resizebox{\textwidth}{!}{%
\begin{tabular}{|c|ccc|cc|ccccc|}
\hline
Architecture & Iterations  & Iterations  & Iterations  & $h$  & $\beta$  & $\alpha_1$  & $\alpha_2$  & $\alpha_3$  & $\alpha_4$  & $\alpha_5$  \\
&  (free phase)  & (clamped phase)  & (loop) &  &  &  &  &  &  &  \\ \hline
$784 - 1500 - 1500 - 10$ & $16$  & $8$  & $2$ & $0.5$ & $1.0$ & $0.2$  & $0.05$  & $0.01$  & $0.0$  &  \\
$784 - 1500 - 1500 - 1500 - 10$ & $20$  & $10$  & $2$  & $0.5$  & $1.0$  & $0.064$  & $0.026$  & $0.004$ & $0.001$ & $0.0$  \\ \hline
\end{tabular}}
\caption{The hyperparameters for different network architectures. $h$ is the time constant of the neuron. $\beta$ is clamping factor for weakly clamped phase. $\alpha$ is the per layer learning rate for parameter updates. The loop iterations is the number of times to loop over the free phase, clamped phase, and update rule per training example. }
\label{hyperparams}
\end{table*}

\section{Experimental Results}
Here we detail experimental results on the performance and behavior of our model described in sections 2 and 3. We provide evidence that our model is trainable and has comparable performance for networks of the same size on the selected datasets. We look at the behavior of our model during training and how the feedback loop drives weight changes. We have kept the learning algorithm bare and selected two similar datasets with a substantive, not enormous, difference in difficulty. Our selection enables us to focus on the mechanism of Feedfoward Propogation and have considerable literature to use as a baseline. The two and three hidden layer architectures detailed in Table \ref{hyperparams} are trained on the MNIST and Fashion MNIST datasets \cite{lecun1998mnist,xiao2017fashion}. The hyperparameters in the networks are fixed and were chosen from several trial runs. The two layer architecture was run for sixty epochs and the three layer for one hundred and fifty epochs. The best model during the entire run was kept.
    All experiments were carried out using Python and implemented in Theano \cite{bergstra2010theano,bastien2012theano}.

\subsection{MNIST and Fashion MNIST datasets}
The MNIST dataset consists of 60,000 training examples and 10,000 test examples of handwritten digits. We split the training set to have a validation set of 10,000 examples. Each example is a $28x28$ grey scale image with a corresponding label $d \in [0,9]$. The MNIST dataset is useful for showing that a model can learn. The size and simplicity of the MNIST dataset make it convenient and quick to test models on. However, the MNIST dataset is considered to be too easy to reliably evaluate the performance of a model \cite{xiao2017fashion,cohen2017emnist}.

The Fashion MNIST dataset is one choice for a more challenging classification task that is a drop-in replacement for the MNIST dataset. The Fashion MNIST dataset consists of 60,000 training examples and 10,000 test examples of clothing articles, instead of digits. The examples, though, have the same format as the MNIST dataset and are $28x28$ grey scale images with corresponding labels $d \in [0,9]$. The Fashion MNIST dataset is a step up from the MNIST dataset. It remains simple enough that complex architectures, learning algorithms, and models are not needed to view progress and soundly measure performance.

\begin{figure}
    \includegraphics[scale=.5,trim = 0 0 0 0,clip]{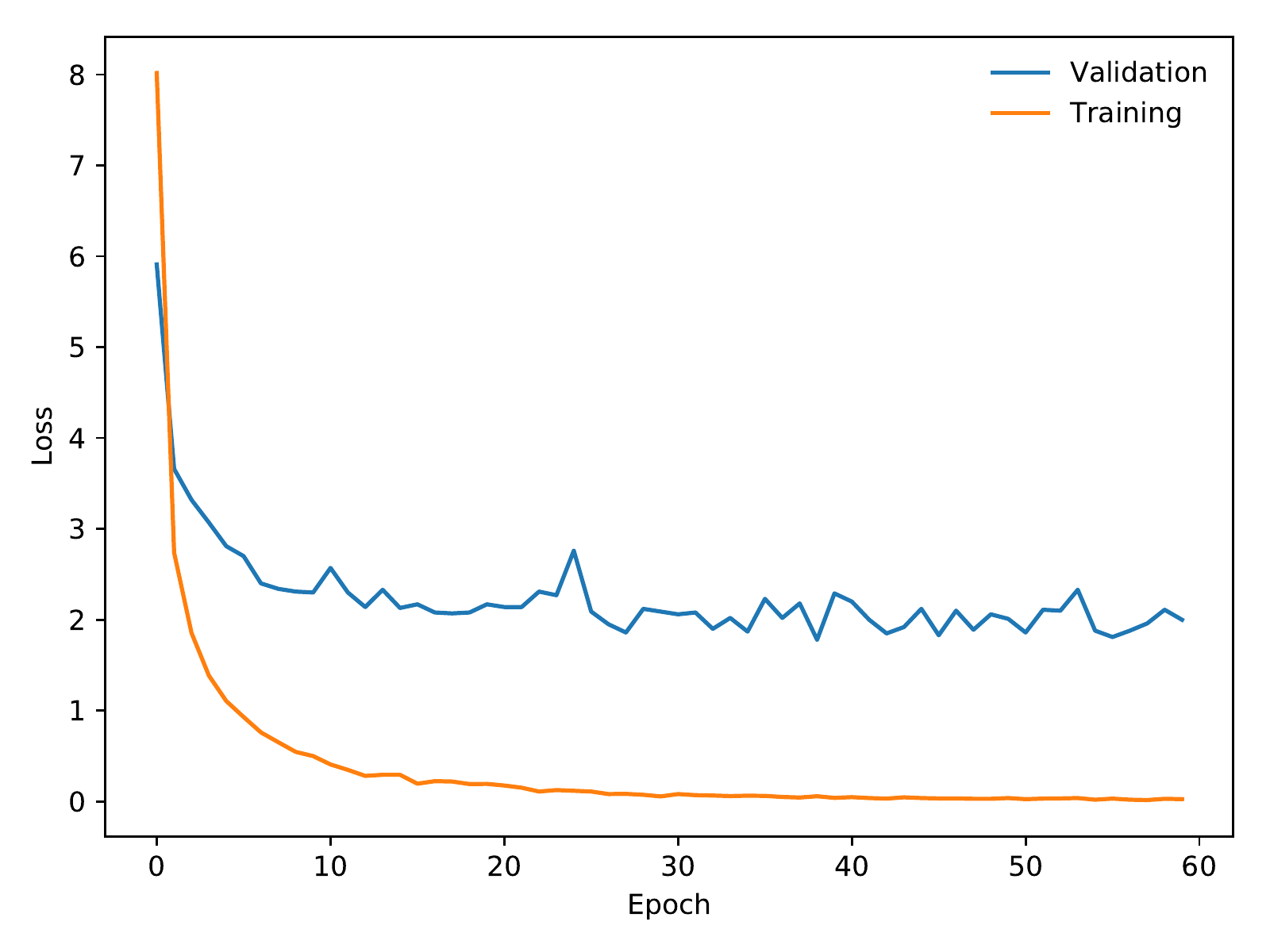}
    \includegraphics[scale=.5,trim = 0 0 0 0,clip]{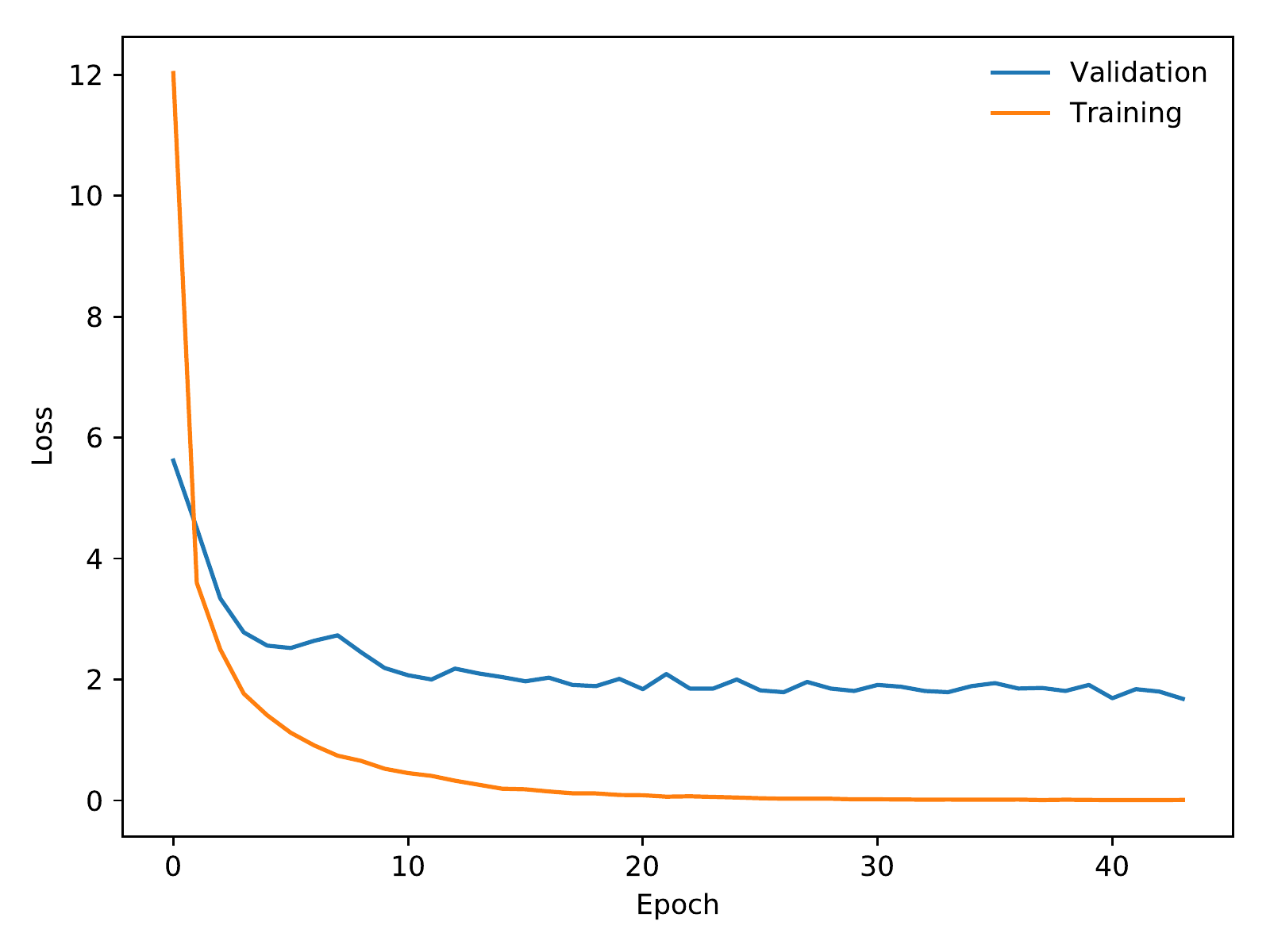}
    \caption{The training and validation error on the MNIST dataset for two neural networks: One with 2 hidden layers of 1500 units (top), and the other with 3 hidden layers of 1500 units (bottom). The two and three layer networks have a validation error between $1.76-1.80\%$ and a test error between $1.85\%-1.90\%$.}
    \label{fig:netmnist}
\end{figure}

\begin{figure}
    \includegraphics[scale=.5,trim = 0 0 0 0,clip]{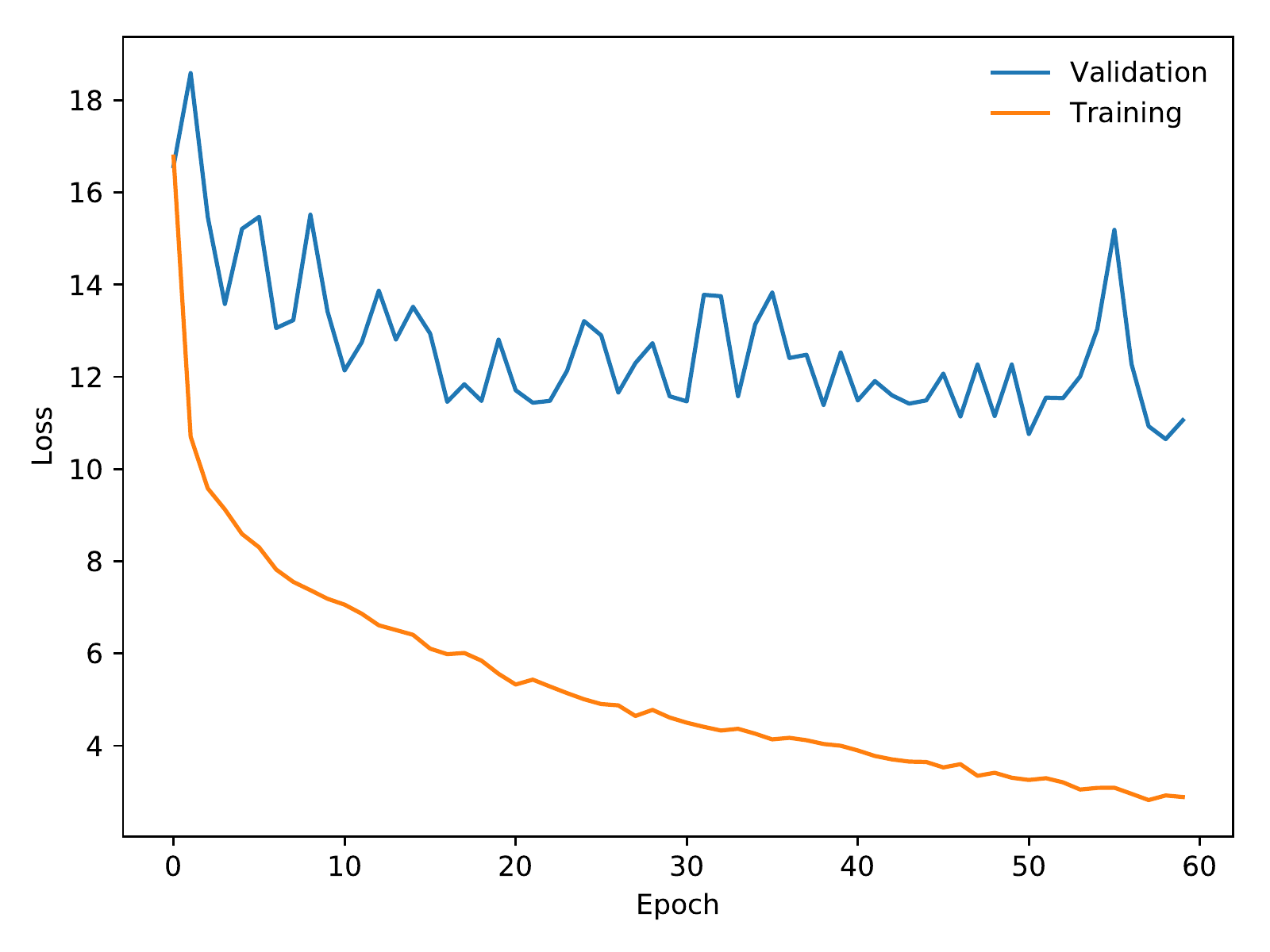}
    \includegraphics[scale=.5,trim = 0 0 0 0,clip]{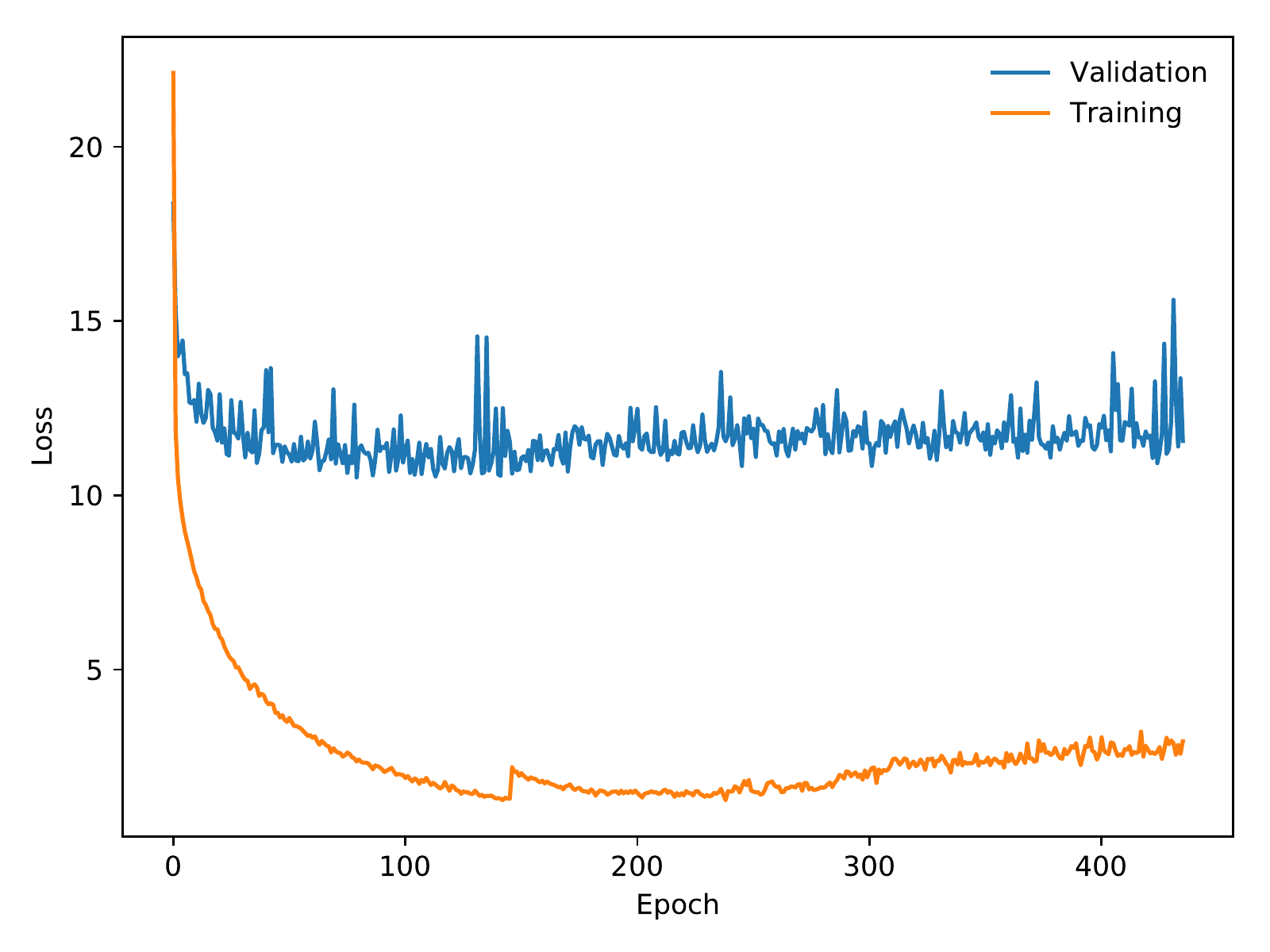}
    \caption{The training and validation error on the Fashion MNIST dataset for two neural networks: One with 2 hidden layers of 1500 units (top) and the other with 3 hidden layers of 1500 units (bottom). The two and three layer networks have a test error of about $11\%$ and the validation error of about $10.95\%$.}
    \label{fig:netfashion}
\end{figure}

On the MNIST dataset, we achieve between a $1.85\%$ to $1.90\%$ generalization error for both the two layer and three layer architectures, figure \ref{fig:netmnist}. The best validation error is between $1.76\%$ and $1.80\%$ and the training error decreases to $0.00\%$. On the Fashion MNIST dataset, we achieve a generalization error of about $11\%$, figure \ref{fig:netfashion}. The best validation error is about $10.95\%$ and the training error decreases to about $2\%$.


\subsection{Lowering the Relaxation Period in the Free Phase}

We explore how holding the network parameters fixed during the free and clamped phases effects the length of the relaxation period. As discussed in section 3, the network parameters are treated as constants in EP learning algorithm, but are not constants in the EP energy function, equation \ref{eq:prototypical}. We posited that holding the network parameters constant is detrimental to the length of the relaxation period. To this end, we proposed updating the network parameters in sequence with the forward and backward phases as part of a larger phase. The network parameters in this scheme are treated more closely as differentiable variables.

The FP update rule, equation \ref{eq:equpdate}, and EP update rule, equation \ref{eq:uniupdate}, were each run in this scheme as part of separate trials. Our experiments show that the lengthy relaxation to a fixed point during the free phase previously found in \cite{scellier2017equilibrium} can be considerably shortened to a small multiple of the network's length. The growth of the relation period in the free phase is linear as compared to the seemingly polynomial growth in the EP learning algorithm. For a network with three hidden layers on the MNIST dataset, a free phase of twenty iterations and a loop of two iterations takes about a fifth of the time per epoch, converges faster, and achieves the same or better performance than with the EP learning algorithm.

\begin{figure*}
    \includegraphics[scale=.45,trim = 0 0 0 0,clip]{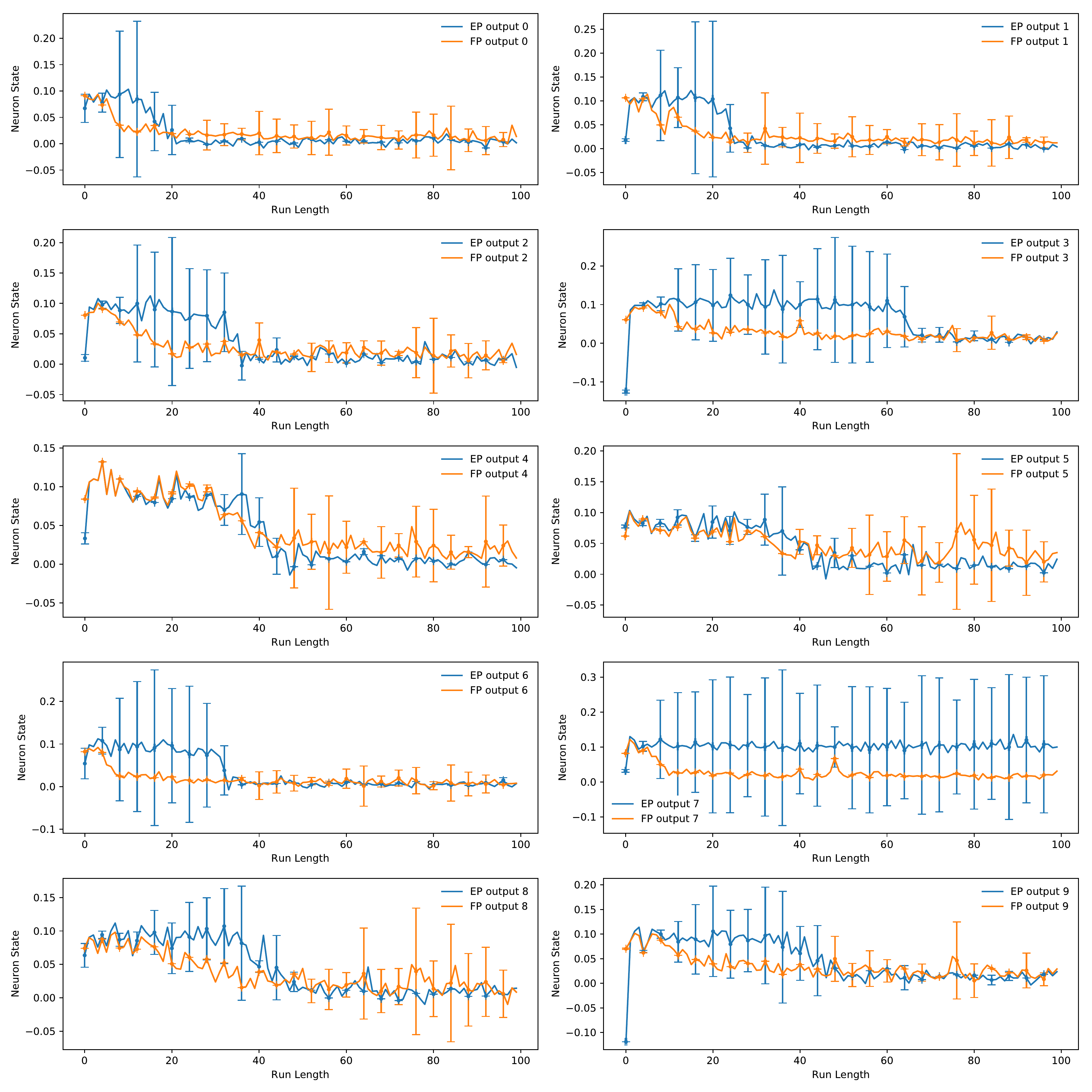}
    \caption{The FP learning algorithm moves toward the optimal target firing rate faster than the EP learning algorithm. The average variance and mean are lower for the FP learning algorithm than for the EP learning algorithm. The blue lines represent a network using the EP learning algorithm. The lines are the mean output error, the mean for the state of the ten output neurons deducted from their target firing rates. The y-axis error bars are the variance difference over batches in a chunk of time. The x-axis error bars are the variance difference over a chunk of time. The variance difference is a comparison between the two lines; the variance for one line has to be larger than the other to show. The target firing rates are the one-hot vector of the targets. The orange lines represent a network using the FP learning algorithm. Both networks use the EP update rule, so the learning algorithms can be comparable. The total run was over 2,500 mini-batches divided in 100 chunks over time. The neurons are averaged over 20 samples per mini-batch. The neurons were simulated for twenty eight steps per batch using the Euler method.}
    \label{fig:outputdiff}
\end{figure*}

The periodic parameter updates were experimentally observed to push the neuron towards the target firing rate without significantly disturbing the trajectory once the neuron was already moving in the direction of the minimum energy state. In figure \ref{fig:outputdiff}, the error of the output neurons over an epoch of the FP learning algorithm (red) and EP learning algorithms (blue) is graphed to show how well the two algorithms meet the target firing rate during training. The FP learning algorithm spends more steps closer to the target firing rate then the EP algorithm. The FP learning algorithm also maintains a lower variance and mean throughout the epoch, on average. These differences are reflected in the training speed and accuracy of the two learning algorithms. For a network with two hidden layers, the FP learning algorithm has a training error of about $8.7\%$ and validation error of $5.5\%$ compared to the $22.9\%$ and $9.3\%$ respectively of the EP learning algorithm.

\subsection{Insights into Error Forward-Propagation}
With the remarkable success of backpropagation, precise symmetric connectivity seemed to be crucial for effective error delivery \cite{rumelhart1986learning}. Random Feedback Weights, however, showed that approximate symmetry is enough and the perceived constraint on deep error propagation shrank \cite{guerguiev2017towards,liao2016important,lillicrap2016random}. This functionally equivalent approximate symmetry is found in Forward Propagation.

\begin{figure*}
    \begin{minipage}[b]{.5\textwidth}
    \includegraphics[scale=.18,trim = 0 0 0 0,clip]{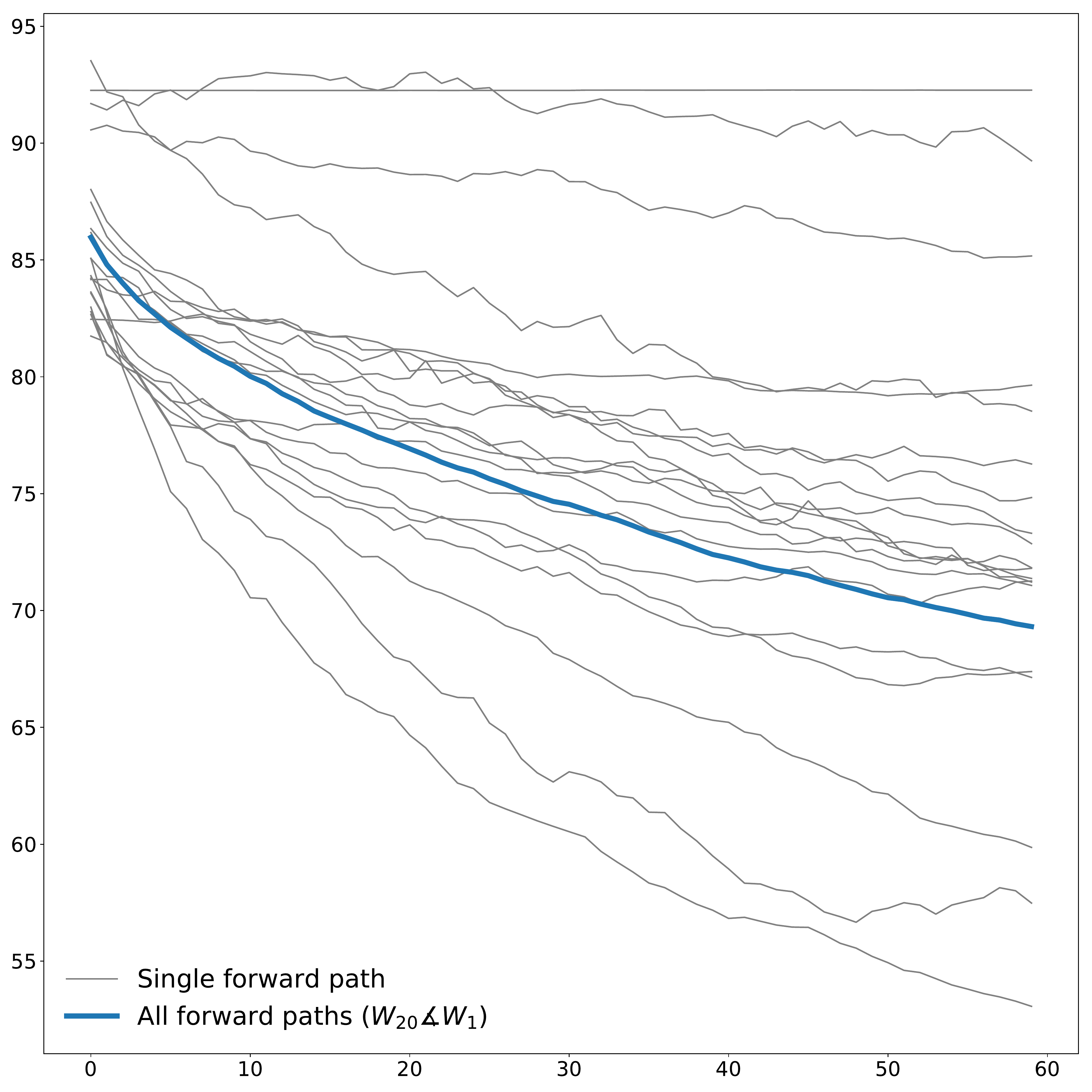}
    \includegraphics[scale=.18,trim = 0 0 0 0,clip]{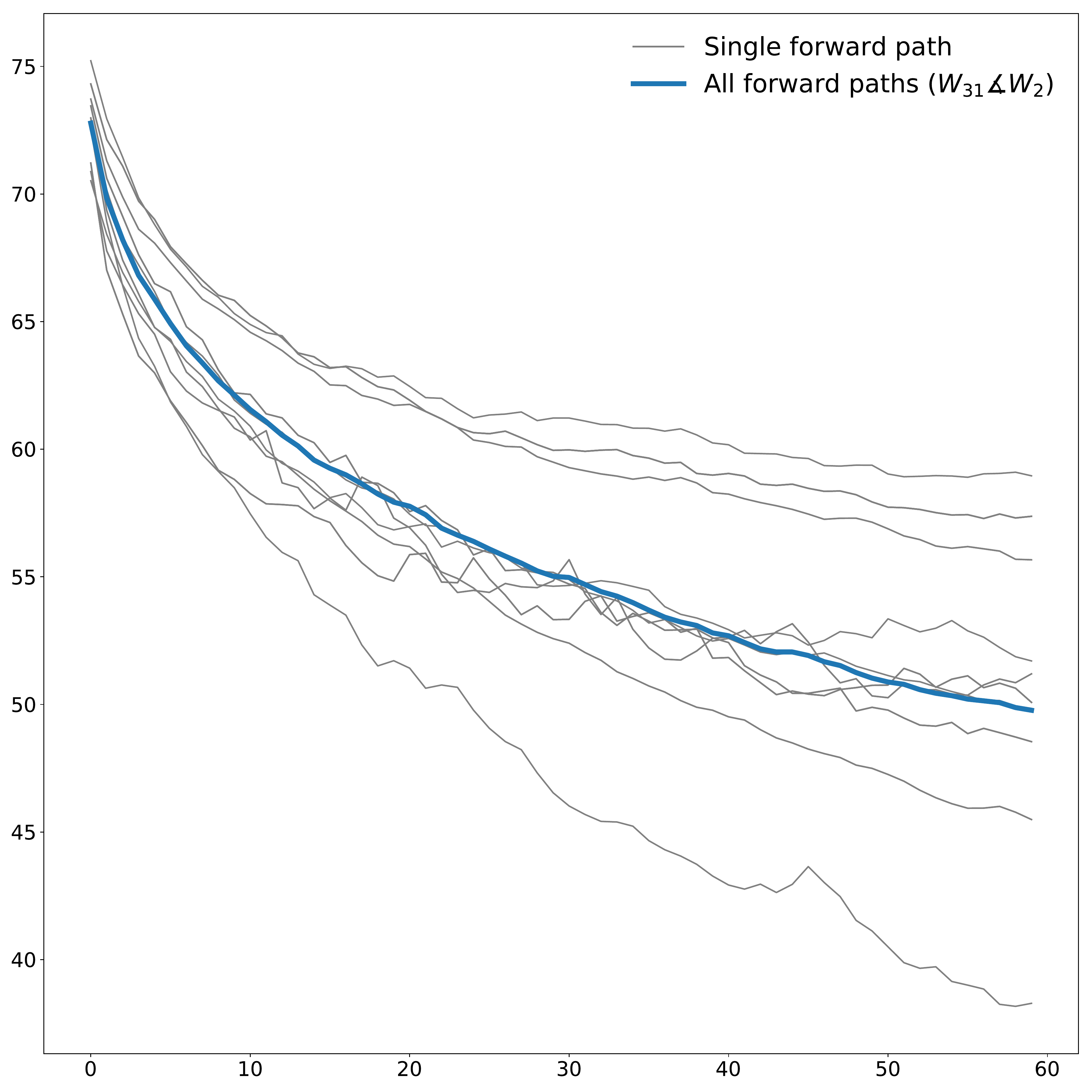}
    \includegraphics[scale=.18,trim = 0 0 0 0,clip]{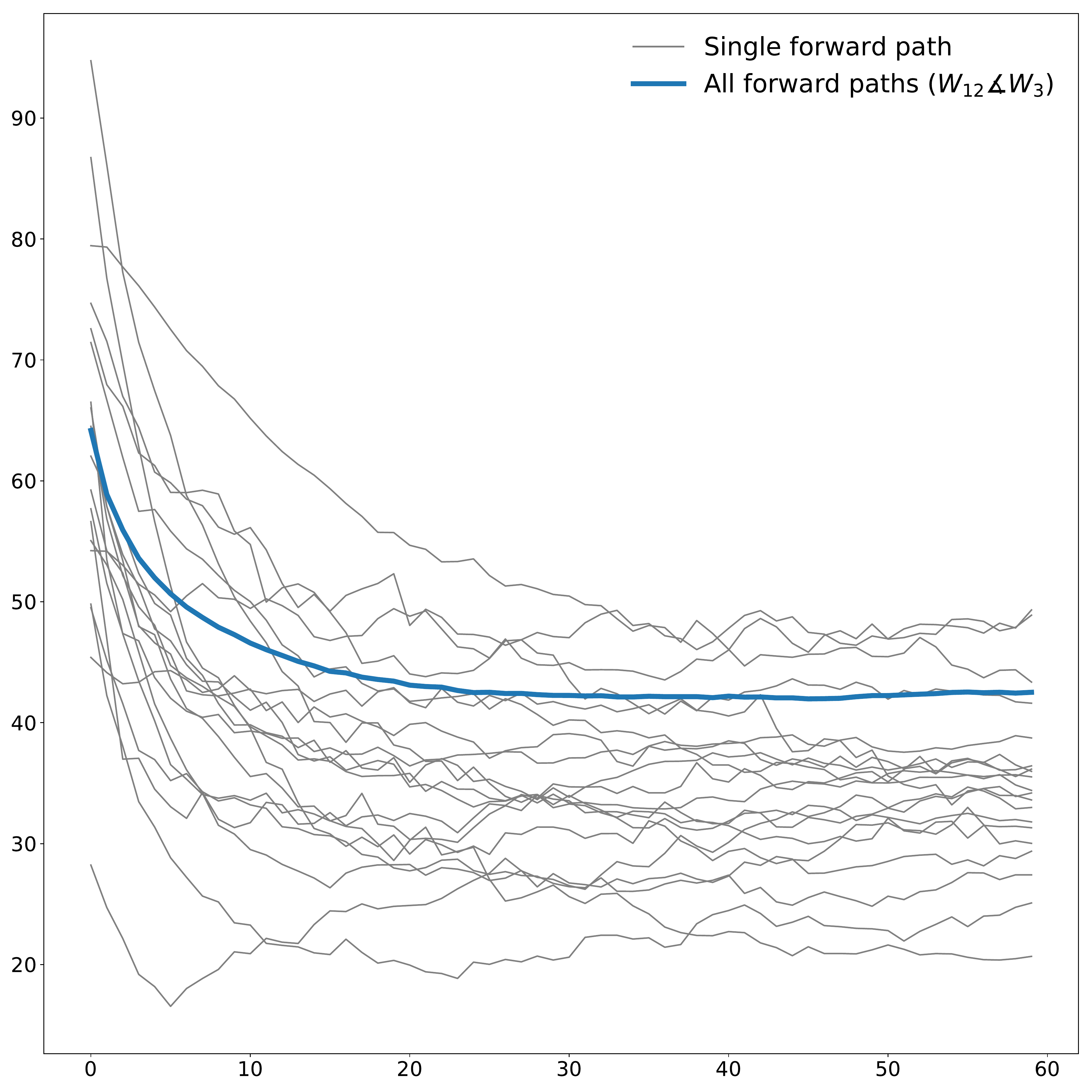}
    \end{minipage}
    \begin{minipage}[b]{.5\textwidth}
    \includegraphics[scale=.18,trim = 0 0 0 0,clip]{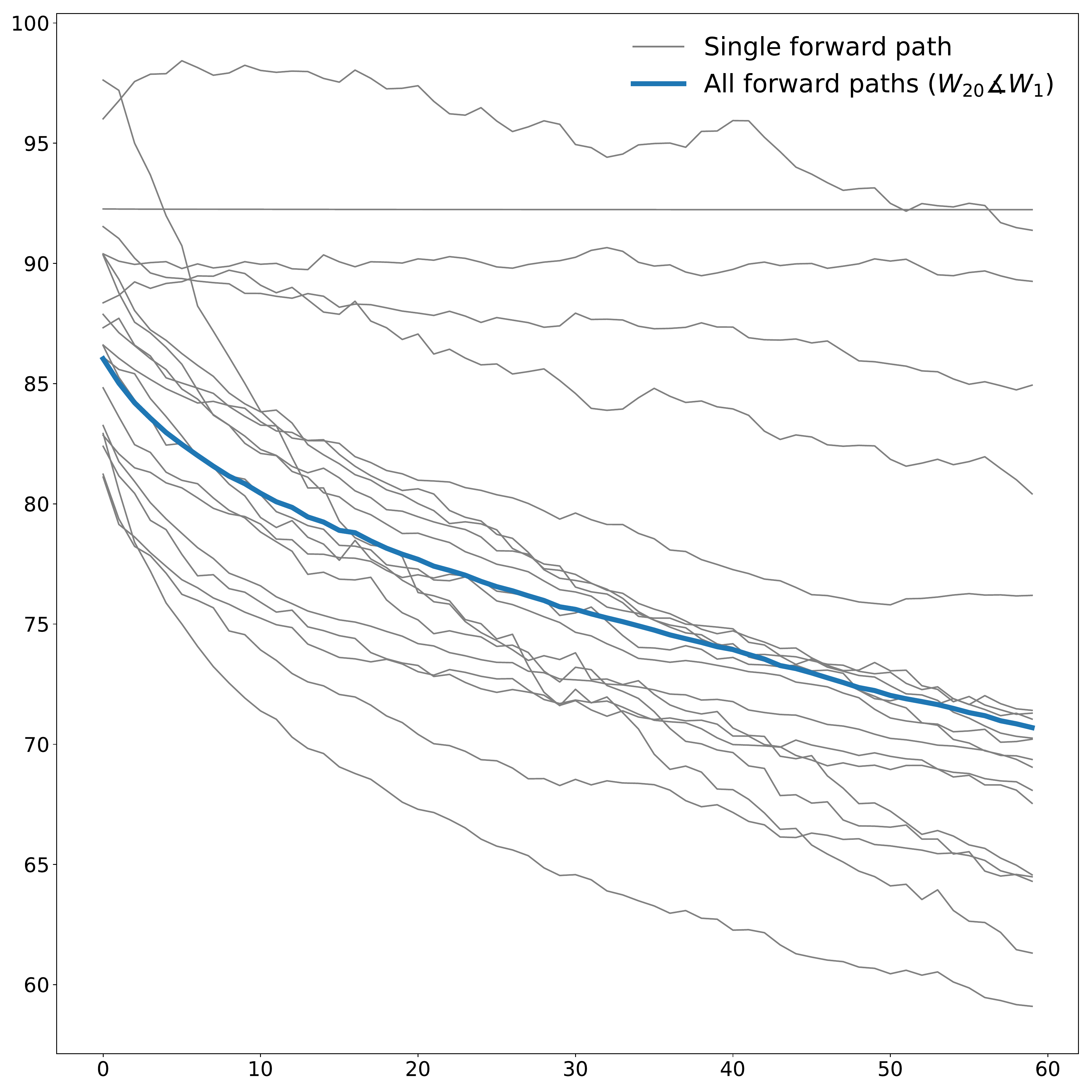}
    \includegraphics[scale=.18,trim = 0 0 0 0,clip]{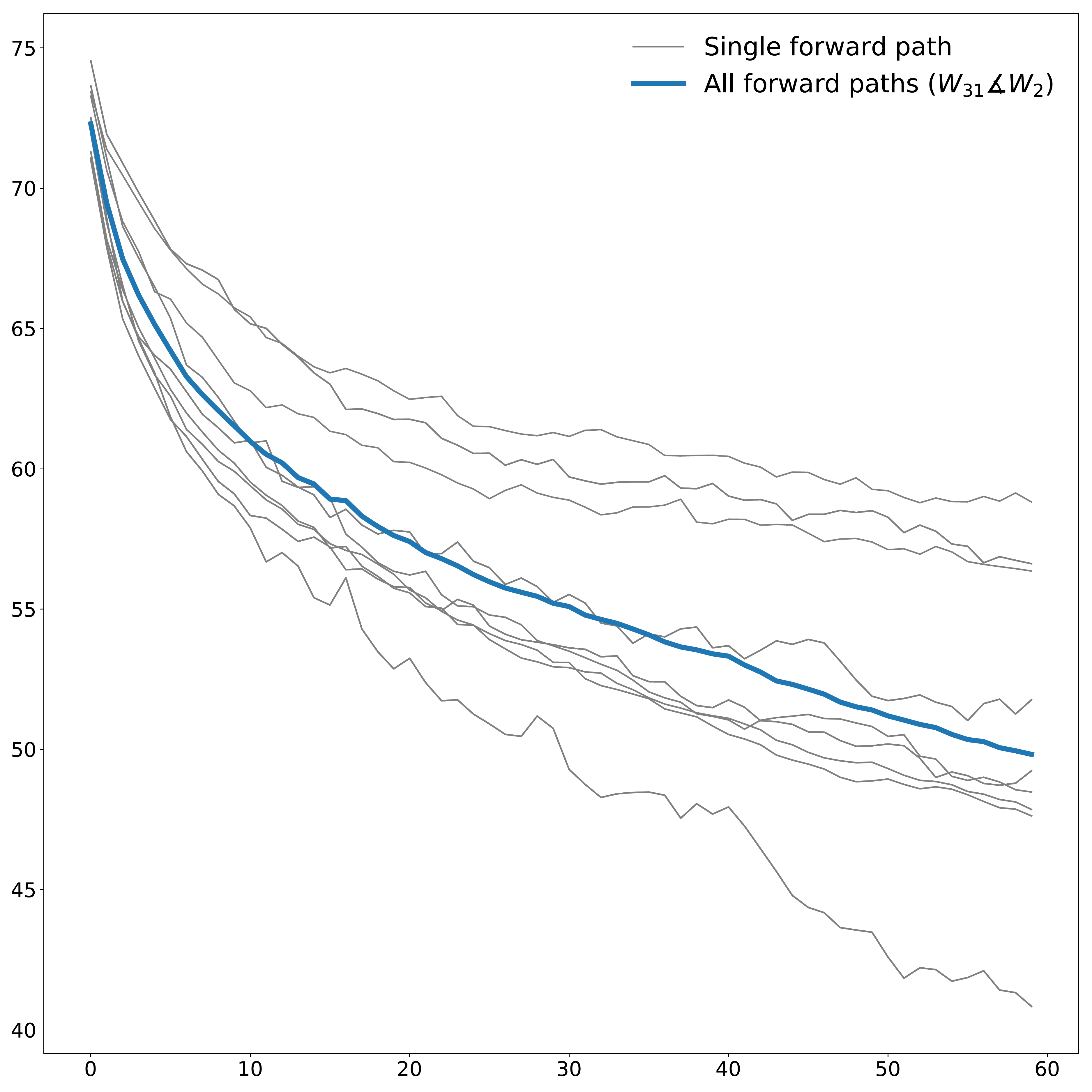}
    \includegraphics[scale=.18,trim = 0 0 0 0,clip]{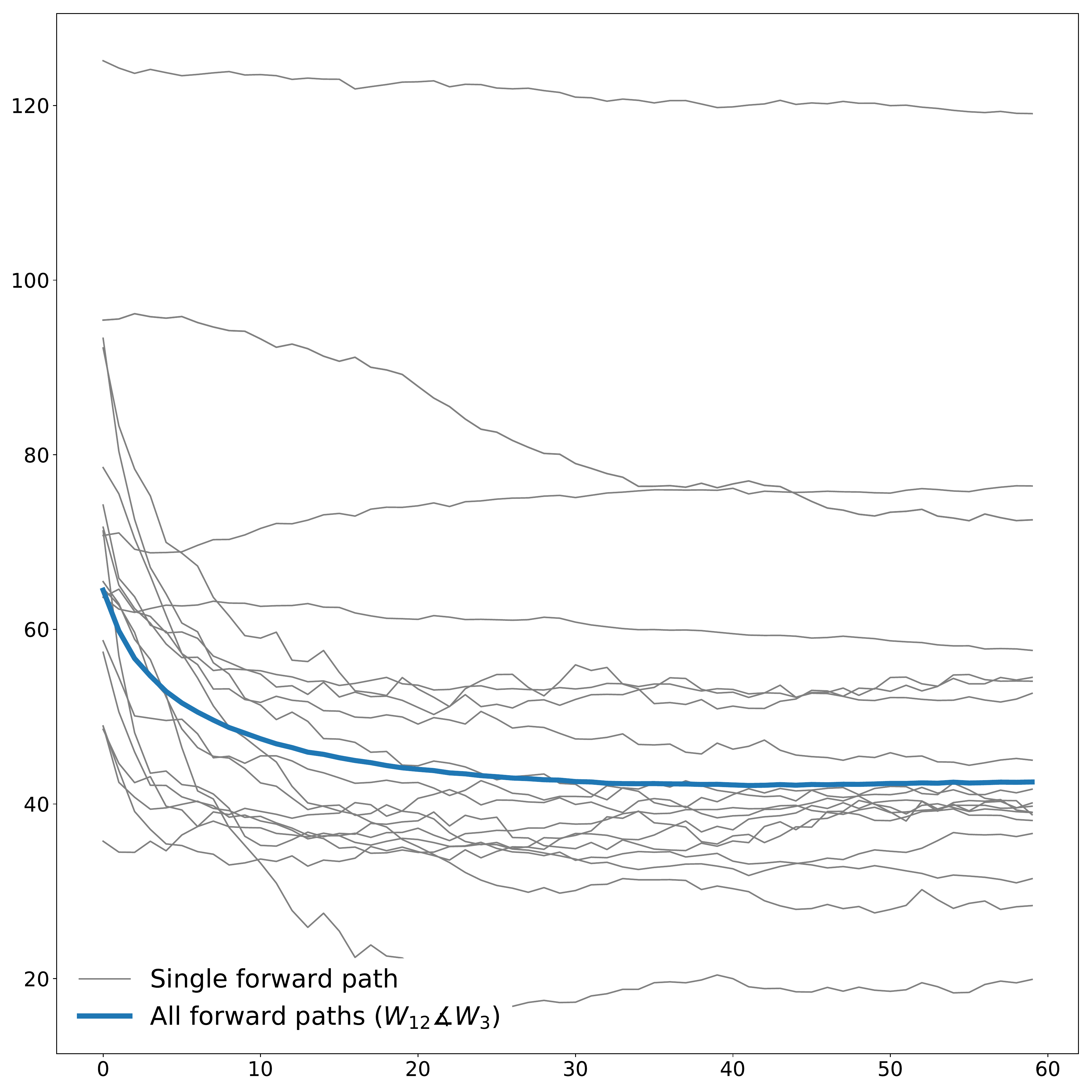}
    \end{minipage}
    \caption{The weight matrices across the network align with each other in a loop. The product of the weight matrices from $x$ to $y$ for $W_{xy}$ align with the weight matrix $z$ for $W_z$ in each graph $W_{xy} \measuredangle W_z$. The alignment occurs when the feedback loop weights are fixed (left) or learned (right). The weight alignment is consistent for individual forward paths from the input receiving neuron to the output neuron (grey) and for all the forward paths (blue).}
    \label{fig:angleloop}
\end{figure*}

With Error Forward-Propagation, the network architecture forms a loop, figure \ref{fig:archefp}. Forward and backward paths only exist in the context of the input and output. Otherwise, the notion of feedback depends on the reference frame of the weight matrix being updated. For a given weight matrix, the feedback weights are all the weights on the path from the downstream error to the presynaptic neuron, which is in general all the other weights in the network loop. The weight matrices in the loop evolve to align with each other \ref{fig:angleloop}. More precisely, each weight matrix aligns with the product of all the other weights in the network loop.

In a network with two hidden layers, let the weights from the input receiving layer back to the output layer be $W_1$ and $W_2$ and weight from the output layer to the input receiving layer be $W_3$. From equation \ref{eq:uniupdate}, $W_2 \propto \rho(\vec{s}_2^0) (\rho(\vec{s}_3^\beta) - \rho(\vec{s}_3^0))$ where $\vec{s}_2 \gets \rho(\vec{s}_1) W_1$, which means information about $W_1$ accumulates in $W_2$. Similarly, $W_1 \propto \rho(\vec{s}_1^0) (\rho(\vec{s}_2^\beta) - \rho(\vec{s}_2^0))$, except since the network architecture is a feedforward loop, $\vec{s}_1 \gets \rho(\vec{s}_3) W_3$, which means information about $W_3$ accumulates in $W_1$. In total, information about $W_3$ and $W_1$ flows into $W_2$ as roughly $W_3 W_1$, which nudges $W_2$ into alignment with the rest of the weights in the loop. This is shown on the right side of figure \ref{fig:angleloop}, where a weight matrix is fixed and the rest of the network's weights come into alignment with the fixed weight. Notice that $W_3 W_1$ has the same shape as $W_2^{T}$ and serves as it's `feedback' weight.

\section{Conclusions, Future Work}
The mechanism underlying the distribution of errors feedback to hidden weights (synapses) and their application in adjusting weights is a principal topic of investigation in neuroscience. The success of deep learning with backpropagation has not been paralleled in neuroscience. Neurons in backpropagation have knowledge of all weights and activations in the forward path to the outputs. Backpropagation multiplies the error signal by the transpose of all weights in the forward path to calculate the loss for each weight. Models using random feedback weights relaxes the requirement of needing to transport all the weights in the forward pathway, while remaining fast and accurate \cite{lillicrap2016random,liao2016important,nokland2016direct,neftci2017event}. Models using random feedback weights ensure the random feedback $\tilde{W}e$ lies within $\angle{90}$ of backpropagation feedback $W^T e$, which generally amounts to having the same sign as the feedforward weights \cite{lillicrap2016random,liao2016important}. Feedforward Propagation, our work, uses Error Forward-Propagation which relaxes the requirements of needing feedback weights, in general, while remaining fast and accurate.

Forward propagation of the feedback is based on four observations. First, regardless of if the learning algorithm is based on backpropagation or random feedback alignment, the error feedback pushes the network in roughly the same direction and nudges the output towards the target. Second, adjusting the input receiving neurons can push the network in any direction just as well as having information transfer through backward connections. Third, the error does not necessarily need to propagate backwards or in any direction as long as the network is pushed roughly in the same direction quickly. Fourth, different phases (or modulators) may be used to separate prediction from learning on the same connections and neurons  \cite{scellier2017equilibrium,xie2003equivalence} instead of separate connections or pathways. The confluence of these observations lead us to reuse the feedforward connections to propagate the error feedback.

In this paper, we practically remove the architectural constraint on connectivity for error feedback in the brain. Only a backward loop connection is required for effective error feedback. The forward connections are reused to deliver the error feedback through hidden layers by feeding the outputs into the input receiving layer. This mechanism, Error Forward-Propagation, is a plausible basis for how error feedback occurs deep in the brain independent of and yet in support of the functionality underlying intricate network architectures.

Future work can focus on how the learning algorithm could be implemented in the brain. Another experiment is to implement Feedforward Propagation as a spiking neural network, much like Equilibrium Propagation \cite{mesnard2016towards}.

\section{Acknowledgements}
This research was partially supported by the Office of Naval Research and we acknowledge the award number N00014-15-1-2126.ONR

{\small
\bibliographystyle{unsrt}
\bibliography{feedforwardenergy}
}

\end{document}